# Teaching Autoregressive Language Models Complex Tasks By Demonstration

**Gabriel Recchia**
Department of Pure Mathematics and Mathematical Statistics
University of Cambridge
`glr29@cam.ac.uk`

## Abstract

This paper demonstrates that by fine-tuning an autoregressive language model (GPT-Neo [1], [2]) on appropriately structured step-by-step demonstrations, it is possible to teach it to execute a mathematical task that has previously proved difficult for Transformers – longhand modulo operations – with a relatively small number of examples. Specifically, we fine-tune GPT-Neo to solve the *numbers__div_remainder* task from the DeepMind Mathematics Dataset; Saxton et al. [3] reported below 40% accuracy on this task with 2 million training examples. We show that after fine-tuning on 200 appropriately structured demonstrations of solving long division problems and reporting the remainders, the smallest available GPT-Neo model achieves over 80% accuracy. This is achieved by constructing an appropriate dataset for fine-tuning, with no changes to the learning algorithm. These results suggest that fine-tuning autoregressive language models on small sets of well-crafted demonstrations may be a useful paradigm for enabling individuals without training in machine learning to coax such models to perform some kinds of complex multi-step tasks.

## 1 Introduction

Models based on the Transformer architecture [4] have achieved impressive results on a wide variety of language-related tasks. They have also shown great promise on non-linguistic tasks such as image recognition [5], image generation [6], navigation [7], and symbolic manipulation [8]. In the mathematical domain, Transformer-based frameworks have been used to solve ordinary differential equations to a higher standard than traditional symbolic algebra systems [9], and can learn to complete certain mathematical exercises such as addition, subtraction, comparison, etc. with sufficient training [3]. Success on mathematical questions is of particular interest, as a system that could accomplish this in a flexible way without being preprogrammed with explicit mathematics knowledge would seem to require several subskills that would be useful for developing systems with more general reasoning and symbolic manipulation capabilities [3]. These include the ability to execute 'sub-algorithms' relevant to solving the problem (e.g., solving a system of equations might also require addition, subtraction, multiplication, etc.), successful determination of which sub-algorithms to execute and when, storage of intermediate values, and perhaps transformation of the input string to a more appropriate representation.

Two caveats are in order about the mathematical achievements of Transformers. The first is that these models typically are trained with very large numbers of training examples. Regarding the two examples of mathematical problem solving mentioned in the previous paragraph, for example, [9] employed training set sizes of 20-40 million examples, and [3] used training sets with 2 million examples per module. This is not necessarily a problem when many examples can be automatically generated or easily acquired, but is not suitable if we want models to learn symbolic manipulation or reasoning tasks for which acquiring training data is costly. The second caveat is that, whereas humans frequently learn complex tasks by imitating others, Transformers are often trained with no reference (or very limited reference) to the sequence of steps that a human would take to solve the task. For example, Saxton *et al.* [3] trained a Transformer on question-answer pairs for over fifty different mathematical tasks. On one hand, the fact that the model achieved over 80% performance for the majority of tasks is remarkable—even given the very large number of training examples—when one considers that the model was given no hints at all about the path from the question to the solution. However, the model achieved scores of less than 40% correct for the five most difficult tasks: calculating remainders under division, two prime factorization tasks, and conversion between bases. Humans can learn how to engage in long division, convert between bases, and compute prime factors with a far smaller number of training examples, but generally learn to do so via observation, often combined with direct instruction. If we wish to take steps toward general reasoning and symbolic manipulation capabilities, approaches that learn by observation from a relatively small number of examples seem like promising ones to pursue.

One contribution of this research is as a practical demonstration of how a small number of explicit demonstrations can serve as an effective training set for tasks involving symbolic manipulation, even when the method of training employed is as simple as fine-tuning an off-the-shelf autoregressive language model. In this paradigm achieving better performance on a task becomes a matter of making tweaks to the training set—a task that can be done without machine learning expertise—rather than making changes to the underlying learning algorithm. In this respect, it is similar to the concept of *data programming* [10], [11], an approach to creating training datasets with numerous imperfect but easy-to-generate 'labeling functions' that has proven popular and successful in industry. Rather than focusing on how to create large quantities of approximately labelled instances, however, the present approach emphasizes the role of finessing the representation of a smaller number of detailed demonstrations of the task at hand.

## 2 Related work

In addition to the work described above, the approach taken in this case study takes inspiration from behavioral cloning [12]–[14], variants of which have been used elsewhere to leverage Transformers for imitation learning [15]–[18]. In behavioral cloning, training data consists of a set of expert demonstrations whereby each demonstration corresponds to a sequence of state/action pairs; given a state, the model's task is to learn to select an action that the expert demonstrator would be likely to select in the same state. Similarly, autoregressive pretrained language models predict the next symbol on the basis of previous symbols, irrespective of what those symbols represent. If such symbols are taken to represent to action and state information, can such models be employed, without further modification beyond fine-tuning, to learn algorithms for manipulating symbolic data in complex ways, from a small number of training examples? Prior literature suggests that this may be possible if the input is represented in forms that facilitate symbolic manipulation. For example, providing full step-by-step solutions to GPT-2 during training improves performance on a dataset of difficult mathematical problems [19], and off-the-shelf pretrained Transformer models achieve near-perfect results on 15-digit addition problems with as few as 1,000 training examples, provided that numbers are notated in a manner that explicitly represents the place value of each digit [20]. Several other investigators have also pre-trained or fine-tuned neural models on synthetic data consisting of mathematical problems [3], [21], [22], although these approaches have generally used very large numbers of training examples. In some more constrained domains, such as reasoning about quantitative comparisons (more than, less than, etc.), even relatively small numbers of synthesized training examples have proven useful [23], [24]. Finally, although we do not condition on rewards or include reward symbols, training autoregressive models on undifferentiated sequences of states and actions is also reminiscent of recent formulations of reinforcement learning as a sequence modeling problem involving undifferentiated sequences of states, actions, and rewards; this setup can also be deployed in an imitation learning context [17], [18] (see also [25]).

Other approaches to constructing training datasets that have proven useful in improving language models' performance on numerical reasoning tasks include the generation of question-answer pairs from structured datasets [26], which are in turn used as training examples. Yet other studies improve performance on mathematical or reasoning tasks expressed in natural language by including components explicitly designed to execute numerical or logical operations [27]–[32]; this is not the approach we take here, but presumably could be combined with more dataset-focused approaches to achieve improved performance on domain-specific tasks. Finally, the use of 'environment forcing' (Section 4.2) is inspired by techniques for minimizing exposure bias [33]–[35] and corresponds to the conditioning on sequences of prior actions and states undertaken by [17], [18] when using 'decision transformers' and 'trajectory transformers' in imitation learning contexts.

## 3 Case study: computing remainders under long division

The following case study demonstrates that by fine-tuning on appropriately represented demonstrations of problem solutions, an off-the-shelf Transformer model can learn to successfully complete the vast majority of long division problems from a random sample of the DeepMind Mathematics Dataset [3] with only 200 training examples while showing its work at every step. This was selected as a case study for two reasons. First, it appears to be challenging for Transformers even when they are provided with vast amounts of training data, as discussed below. Second, long division is known to be a particularly difficult task for young learners, as it requires them to draw upon all or nearly all the arithmetic knowledge that they have acquired up to that point and execute a novel algorithm with many steps [36]. In a French study of fifty-six children aged 10-11, most of whom had started to learn the procedure for long division one year ago, the percentage of correct answers for each problem having a dividend at least two digits long ranged from 21% to 73%; the number of computational steps required was particularly predictive of problem difficulty [36].

Training, validation, and test data were taken from the DeepMind Mathematics Dataset's interpolated *div_remainder* module, which consists of instructions to calculate the remainder under division (e.g., "Calculate the remainder when 25736 is divided by 144"); numbers in these statements range from 1 to 8 digits in length. Saxton et al. [3] demonstrated that a 30M-parameter Transformer model could achieve just under 40% performance on this task after training on $2 \times 10^6$ training examples. In a study of the relationship between scaling of model size and compute budgets, Henighan et al. [21] generated training data for each task in [3] procedurally in an online setting using the same algorithm that had been used to generate their test data. They reported approximately 60% accuracy by a 400M-parameter model on the remainder task, although they acknowledged that the procedurally generated training set may have contained items also in the test set, as the number of training examples was large and overlaps with the test set were not examined.

Given that larger models and additional data continue to improve Transformer models' performance on tasks involving symbolic manipulation, why bother going to the effort of providing them with detailed demonstrations of how to complete particular types of problems? First, if doing so enables learning after a relatively small number of training examples, this could be a promising training approach in circumstances where training data is expensive to create, or in which computational resources are limited. Second, some tasks may be difficult enough to learn that providing demonstrations may enable improvements in performance which would not have been achieved otherwise. Third, demonstrating how to solve a problem step-by-step may permit Transformers to do the same at inference, making the model's process more transparent and highlighting failure points that could be addressed by tweaking subsequent demonstrations. Finally, detailed demonstrations of tasks can be provided by any individual capable of completing the task; expertise with deep learning architectures is not required. In the same way that manual labelling of images by volunteers enabled advances in supervised image classification, so too might manual demonstrations of complex, multi-step tasks involving symbol manipulation enable advances in the abilities of Transformer-based systems to perform such tasks.

## 4    Method

### 4.1 Training

Each training instance corresponded to a problem from the DeepMind Mathematics Dataset, followed by a demonstration of how to complete the problem represented as a sequence of symbols (see sec. 4.1.1.). The input question (e.g., "What is the remainder when 203462 is divided by 591?") was represented in a format that explicitly labeled the position of each character within each word, in order to circumvent the drawbacks of the default position embeddings and tokenizer [20]. This is similar to the approach taken by Nogueira, Jiang and Lin [20], but without any explicit indication of place value.

200 such training examples were used to fine-tune 125M-parameter GPT-Neo models for between 5,000 and 15,000 steps using aitextgen version 0.5.1[37] and a batch size of 2, which permitted all training to take place on a single NVIDIA Tesla P100. The default AdamW optimizer was used with a learning rate of $5 \times 10^{-5}$.

Problems for training, validation, and for the first set of test questions were sampled from the pre-generated files associated with the *numbers__div_remainder* task in the DeepMind Mathematics Dataset [3] with an equal number sampled uniformly from each of the *train-easy*, *train-medium*, and *train-hard* datasets, the approach that had been used to generate training data in [3] (see https://github.com/deepmind/mathematics_dataset#readme). The second test set of 500 examples was sampled randomly from the pre-generated 'interpolated' data for *numbers__div_remainder*, the test set employed in [3]. No input questions were duplicated among the training, validation or test sets.

### 4.1.1 Representation of the demonstrations used in training

The state of the environment was represented as a 99 x 99 grid, each square of which can contain a symbol (Figure 1). For each problem sampled for training, a corresponding demonstration of how to solve it was generated by the "demonstrator", a script available at https://github.com/mesotron/teaching_transformers/. However, the structure of these demonstrations was such that the "demonstrators" could in principle have been individuals manually pointing/clicking/typing on an appropriate graphical interface. Specifically, the actions available to the demonstrator were as follows:

- to *write* a symbol (*s*) at an x-y coordinate (*coord*), represented by a string of the form "write *coord s*";
- to *look* at a particular coordinate on the grid, represented by a string of the form "look *coord s*", where *s* represents the symbol observed at coordinate *coord*;
- to *clear* the rightmost two-fifths of the grid (allowing it to be used as replaceable 'scratch paper'), represented by the single token "clear";
- a *no-op* represented by an arbitrary string within curly braces, which has no effect upon the grid and returns no state information; the utility of no-ops is discussed in more detail in the Discussion and the caption of Figure 2. A no-op of the form "{ final remainder is $X$ }" always appears at the end of the demonstration.

**Figure 1.** Example state of the environment after a complete demonstration. Slashes represent cases where the demonstrator has (purposefully) overwritten a symbol at one location with a new symbol.

|    | 00 | 01 | 02 | 03 | 04 | 05 | 06 | 07 | 08 | 09 | 10 | 11 | 12 | 13 | 14 | 15 | ... | 60 | 61 | 62 | 63 | 64 | 65 | 66 | 67 | 68 | 69 | 70 |
|----|----|----|----|----|----|----|----|----|----|----|----|----|----|----|----|----|-----|----|----|----|----|----|----|----|----|----|----|----|
| 00 |    |    |    |    |    |    |    |    |    |    |    |    |    |    |    |    |     |    |    |    |    |    |    |    |    |    |    |    |
| 01 |    |    |    |    |    | 1  | 1  | 6  |    |    |    |    |    |    |    |    |     |    |    |    |    |    |    |    | +  | 1  | 6  |    |
| 02 | 1  | 6  |    | 1  | 8  | 6  | 2  |    |    |    |    |    |    |    |    |    |     |    |    |    |    |    |    |    |    | 1  | 6  |    |
| 03 |    |    | −  | 1  | 6  |    |    |    |    |    |    |    |    |    |    |    |     |    |    |    |    |    |    |    | +  | 1  | 6  |    |
| 04 |    |    |    |    | 2  | 6  |    |    |    |    |    |    |    |    |    |    |     |    |    |    |    |    |    |    |    | 3  | 2  |    |
| 05 |    |    |    | −  | 1  | 6  |    |    |    |    |    |    |    |    |    |    |     |    |    |    |    |    |    |    | +  | 1  | 6  |    |
| 06 |    |    |    |    | 1̸0̸ | 0̸ | 2  |    |    |    |    |    |    |    |    |    |     |    |    |    |    |    |    |    |    | 4  | 8  |    |
| 07 |    |    |    | −  |    | 9  | 6  |    |    |    |    |    |    |    |    |    |     |    |    |    |    |    |    |    | +  | 1  | 6  |    |
| 08 |    |    |    |    |    | 6  |    |    |    |    |    |    |    |    |    |    |     |    |    |    |    |    |    |    |    | 6  | 4  |    |
| 09 |    |    |    |    |    |    |    |    |    |    |    |    |    |    |    |    |     |    |    |    |    |    |    |    | +  | 1  | 6  |    |
| 10 |    |    |    |    |    |    |    |    |    |    |    |    |    |    |    |    |     |    |    |    |    |    |    |    |    | 8  | 0  |    |
| 11 |    |    |    |    |    |    |    |    |    |    |    |    |    |    |    |    |     |    |    |    |    |    |    |    | +  | 1  | 6  |    |
| 12 |    |    |    |    |    |    |    |    |    |    |    |    |    |    |    |    |     |    |    |    |    |    |    |    |    | 9  | 6  |    |
| 13 |    |    |    |    |    |    |    |    |    |    |    |    |    |    |    |    |     |    |    |    |    |    |    |    | +  | 1  | 6  |    |
| 14 |    |    |    |    |    |    |    |    |    |    |    |    |    |    |    |    |     |    |    |    |    |    |    |    | 1  | 1  | 2  |    |
| 15 |    |    |    |    |    |    |    |    |    |    |    |    |    |    |    |    |     |    |    |    |    |    |    |    |    |    |    |    |

For example, applying the sequence of actions represented by the demonstration whose beginning is illustrated in Figure 2 would result in the state of the environment represented in Figure 1. The entire demonstration is given in Figure 3. Multiple coordinates and symbols are permitted to follow a single *write* or *look* token; i.e., write $coord_1$ $s_1$ $coord_2$ $s_2$ was treated as shorthand for write $coord_1$ $s_1$ write $coord_2$ $s_2$. Because demonstrations are merely representations of a series of steps intended to make it easier for the model to learn to arrive at the desired answer and are not provided at test time, information within them can be represented in whatever form is deemed likeliest to facilitate learning and generalization to other problems of the same kind. For example, in Figures 2 & 3, x-y coordinates are represented in the format *x,y:consecutive_symbol_count*, where the term after the colon is a unique token that corresponds to the number of consecutive symbols that have appeared in a row since the last empty grid space[1]. GPT-Neo's default tokenizer was used, and the representation chosen for the demonstrations was informed by its byte pair encoding vocabulary. For example, the tokenizer treats all single-digit numerals, double-digit numerals, and three-digit numerals beginning with '2' as unique and consistent tokens so long as they do not appear immediately adjacent to any other numerals. As a convention, therefore, single-digit numerals were used within demonstrations to represent symbols or values, two-digit numerals to represent coordinates, and three-digit numerals beginning with '2' to represent consecutive symbol counts.

The number of iterations that the model was trained for and the effects of four different approaches to representing demonstrations were explored on a validation set of 100 instances. The four approaches explored were: (1) training data that included strings corresponding to *write* actions, *look* actions, and no-ops, as in Figure 3; (2) *write* and *look* actions only (no no-ops); (3) *write* actions only; and (4) no actions at all, only the prompt followed by the desired answer. For each approach, we identified the model that maximized performance on a validation set of 100 examples (Table 1), and final evaluations were conducted on the two test sets of 500 examples each previously described.

---

[1] Grid spaces not containing 'word characters' (A-Z, a-z, 0-9) were treated as empty for purposes of computing consecutive symbol counts.

**Figure 2.** The rightmost column illustrates the beginning of an demonstration representing the sequence of actions that gave rise to the state depicted in Figure 1. Training instances consisted of an input question (represented as described in section 2.1; see example at beginning of Figure 3) immediately followed by a corresponding demonstration. At test time input questions only are shown, leaving the generation of the demonstration to the model. Note that strings inside of curly braces are no-ops (i.e., they do not correspond to any action) but can be included by the demonstrator in hopes of improving learnability. For example, suppose we would like the model to learn to look at a number A located at $coord_1$, then at a number B located at $coord_2$, and finally to write some symbol C if B is greater than A, and to write D if B is smaller than A. We could attempt this by ensuring the training data has some examples of the form "look $coord_1$ A $coord_2$ B write $coord_3$ C" (where B > A), and some examples of the form "look $coord_1$ A $coord_2$ B write $coord_3$ D" (where B < A), and hope that the relationship is learnt. However, by inserting some redundant information such as "look $coord_1$ A $coord_2$ B { A , B larger } write $coord_3$ C" (where B > A) and "look $coord_1$ A $coord_2$ B { A , B smaller } write $coord_3$ D" (where B < A), we might hope to better facilitate the learning of the relationship. The intuition is (1) learning that "{ A , B " consistently follows "look $coord_1$ A $coord_2$ B" is a very easy relationship to learn; (2) learning the circumstances in which *larger* or *smaller* should follow "look $coord_1$ A $coord_2$ B { A , B" may be easier than directly learning the circumstances in which C or D should follow "look $coord_1$ A $coord_2$ B write $coord_3$", as the most relevant symbols (*A* and *B*) have been made more redundant/salient in the former; and (3) learning that *C* tends to follow "*larger }* write $coord_3$" whereas *D* tends to follow "*smaller }* write $coord_3$" should be a very easy relationship to learn. Furthermore, if there are other contexts in which subsequent actions depend on the relative magnitude of two symbols, then including no-ops of the form "{ X , Y larger }" or "{ X , Y smaller }" in those contexts may help to direct attention accordingly. In other words, no-ops of this kind may serve as miniature curricula that make complex relationships easier to learn; they may achieve this by making relevant tokens more salient, and by making crucial steps of multi-step problems explicit rather than implicit.

| Sequence of actions we would like to demonstrate | String representation |
|---|---|
| The demonstrator writes out the division problem "divide 16 into 1862" (1862 divided by 16), using f to represent the long division symbol ⌐. This will entail writing the first symbol of the problem at (0,2), the second symbol at (1, 2), etc. | `write 00,02:201 1 01,02:202 6 02,02:203 f 03,02:201 1 04,02:202 8 05,02:203 6 06,02:204 2` |
| The demonstrator indicates that we are about to enter a context in which we will be comparing two numbers. (Specifically, we are about to determine whether the (n-digit) divisor is smaller than the first n digits of the dividend.) | `{ compare }` |
| The demonstrator looks at the divisor, beginning by looking at grid space (0, 2), where a "1" is observed. The demonstrator continues looking at each symbol to the right until reaching the end of the divisor. | `look 00,02:201 1 01,02:202 6 02,02:203 f` |
| The demonstrator indicates that the divisor had two digits. | `{ 2 digits }` |
| The demonstrator looks at the next n digits of the dividend (where n is the number of digits that were in the divisor). | `look 03,02:201 1 04,02:202 8` |
| The demonstrator starts to compare the numbers. The demonstrator observes that the numbers each have the same number of digits (2), and therefore needs to look at each number's first digit to determine which number is larger. It is observed that their first digits are equal (both are 1), so the demonstrator takes a look at the second digits. The second digit of the first number is 6, and the second digit of the second number is 8, which is larger. So the second number (18) is larger than the first (16). | `{ 2 digits equal } look 71,02:201 1 look 03,02:201 1 { 1 , 1 equal } look 72,02:202 6 look 04,02:202 8 { 6 , 8 larger }` |
| This means that 16 goes into 18 at least once, but the demonstrator will need to figure out exactly how many times 16 goes into 18. The demonstrator kicks things off by writing a zero at the beginning of the quotient as a placeholder, and clears off the scratch paper. | `write 04,01:201 0 clear` |
| The demonstrator looks at the divisor, which is 16, and writes the addition problem "0 + 16" to the scratch paper. In subsequent steps, the demonstrator will keep on adding 16 until reaching a number that's larger than the second number (18); this will reveal how many times the divisor goes into the second number. | `look 00,02:201 1 01,02:202 6 write 71,00:201 _ 72,00:201 0 70,01:201 + 71,01:201̄ 1 72,01:202 6` |
| The demonstrator begins this addition problem, starting in the ones place column. The demonstrator looks at the 0, looks at the 6, and notes that 0 + 6 equals 6. The demonstrator also makes note of the locations of the relevant 0 and the 6 on the grid, and writes the answer underneath. | `look 72,00:201 0 72,01:202 6 { 72,00:201 0 + 72,01:202 6 = 6 } write 72,02:201 6` |
| The demonstrator moves on to the tens place column of the addition problem. There is a blank square on top, and a 1 underneath. The demonstrator notes that nothing (a blank square) plus 1 equals 1, and writes the answer underneath. | `look 71,00:201 _ 71,01:201 1 { 71,00:201 _ + 71̄,01:201 1 = 1 } write 71,02̄:201 1` |

**Figure 3.** The complete demonstration that results in the final state of the environment illustrated in Figure 1. During training, the model is trained on two hundred strings of this kind, separated by end-of-document markers. During validation and testing, it is presented with novel question prompts (e.g., the text in blue), and generates continuations (the text in black). The substring "{ final remainder is ___ }" it eventually generates is inspected to evaluate whether it solved the problem correctly.

```
201 W 202 h 203 a 204 t 200 _ 201 i 202 s 200 _ 201 t 202 h 203 e 200 _ 201 r 202 e 203 m 204 a 205 i 206 n
207 d 208 e 209 r 200 _ 201 w 202 h 203 e 204 n 200 _ 201 1 202 8 203 6 204 2 200 _ 201 i 202 s 200 _ 201 d
202 i 203 v 204 i 205 d 206 e 207 d 200 _ 201 b 202 y 200 _ 201 1 202 6 203 ? 200 _ | write 00,02:201 1
01,02:202 6 02,02:203 f 03,02:201 1 04,02:202 8 05,02:203 6 06,02:204 2 { compare } look 00,02:201 1 01,02:202
6 02,02:203 f { 2 digits } look 03,02:201 1 04,02:202 8 { 2 digits equal } look 00,02:201 1 look 03,02:201 1
{ 1 , 1 equal } look 01,02:202 6 look 04,02:202 8 { 6 , 8 larger } write 04,01:201 0 clear look 00,02:201 1
01,02:202 6 write 71,00:201 _ 72,00:201 0 70,01:201 + 71,01:201 1 72,01:202 6 look 72,00:201 0 72,01:202 6 {
72,00:201 0 + 72,01:202 6 = 6 } write 72,02:201 6 look 71,00:201 _ 71,01:201 1 { 71,00:201 _ + 71,01:201 1 =
1 } write 71,02:201 1 look 70,00:201 _ 70,01:201 + { 0 + 0 = 0 } { compare } look 70,02:201 _ 71,02:201 1
72,02:202 6 73,02:203 _ { 2 digits } look 03,02:201 1 04,02:202 8 { 2 digits equal } look 71,02:201 1 look
03,02:201 1 { 1 , 1 equal } look 72,02:202 6 look 04,02:202 8 { 6 , 8 larger } look 04,01:201 0 { 04,01:201
0 + 1 = 1 } write 04,01:201 1 70,03:201 + 71,03:201 1 72,03:202 6 look 72,02:202 6 72,03:202 6 { 72,02:202 6
+ 72,03:202 6 = 1 2 } write 72,04:201 2 { carry the 1 } look 71,02:201 1 71,03:201 1 { 71,02:201 1 + 71,03:201
1 = 2 } { 2 + 1 = 3 } write 71,04:201 3 look 70,02:201 _ 70,03:201 + { 0 + 0 = 0 } { compare } look 70,04:201
_ 71,04:201 3 72,04:202 2 73,04:203 _ { 2 digits } look 03,02:201 1 04,02:202 8 { 2 digits equal } look
71,04:201 3 look 03,02:201 1 { 3 , 1 smaller } look 70,02:201 _ 71,02:201 1 72,02:202 6 73,02:203 _ write
02,03:201 - write 03,03:201 1 04,03:202 6 look 04,02:202 8 look 04,03:202 6 { 8 , 6 smaller } { 8 - 6 = 2 }
write 04,04:201 2 look 03,02:201 1 look 03,03:201 1 { 1 , 1 equal } { 1 - 1 = 0 } write 03,04:201 0 look
03,04:201 0 write 03,04:201 _ look 04,04:201 2 { read the answer } look 04,04:201 2 05,04:202 _ look 05,02:203
6 write 05,04:202 6 write 05,01:202 0 clear look 00,02:201 1 01,02:202 6 write 71,00:201 _ 72,00:201 0
70,01:201 + 71,01:201 1 72,01:202 6 look 72,00:201 0 72,01:202 6 { 72,00:201 0 + 72,01:202 6 = 6 } write
72,02:202 6 look 71,00:201 _ 71,01:201 1 { 71,00:201 _ + 71,01:201 1 = 1 } write 71,02:201 1 look 70,00:201
_ 70,01:201 + { 0 + 0 = 0 } { compare } look 70,02:201 _ 71,02:201 1 72,02:202 6 73,02:203 _ { 2 digits }
look 04,04:201 2 05,04:202 6 06,04:203 _ { 2 digits equal } look 71,02:201 1 look 04,04:201 2 { 1 , 2 larger
} look 05,01:202 0 { 05,01:202 0 + 1 = 1 } write 05,01:202 1 70,03:201 + 71,03:201 1 72,03:202 6 look
72,02:202 6 72,03:202 6 { 72,02:202 6 + 72,03:202 6 = 1 2 } write 72,04:202 2 { carry the 1 } look 71,02:201
1 71,03:201 1 { 71,02:201 1 + 71,03:201 1 = 2 } { 2 + 1 = 3 } write 71,04:201 3 look 70,02:201 _ 70,03:201 +
{ 0 + 0 = 0 } { compare } look 70,04:201 _ 71,04:201 3 72,04:202 2 73,04:203 _ { 2 digits } look 04,04:201 2
05,04:202 6 06,04:203 _ { 2 digits equal } look 71,04:201 3 look 04,04:201 2 { 3 , 2 smaller } look 70,02:201
_ 71,02:201 1 72,02:202 6 73,02:203 _ write 03,05:201 - write 04,05:201 1 05,05:202 6 look 05,04:202 6 look
05,05:202 6 { 6 , 6 equal } { 6 - 6 = 0 } write 05,06:201 0 look 04,04:201 2 look 04,05:201 1 { 2 , 1 smaller
} { 2 - 1 = 1 } write 04,06:201 1 look 04,06:201 1 { read the answer } look 04,06:201 1 05,06:202 0 06,06:203
_ look 06,02:204 2 write 06,06:203 2 write 06,01:203 0 clear look 00,02:201 1 01,02:202 6 write 71,00:201 _
72,00:201 0 70,01:201 + 71,01:201 1 72,01:202 6 look 72,00:201 0 72,01:202 6 { 72,00:201 0 + 72,01:202 6 = 6
} write 72,02:202 6 look 71,00:201 _ 71,01:201 1 { 71,00:201 _ + 71,01:201 1 = 1 } write 71,02:201 1 look
70,00:201 _ 70,01:201 + { 0 + 0 = 0 } { compare } look 70,02:201 _ 71,02:201 1 72,02:202 6 73,02:203 _ { 2
digits } look 04,06:201 1 05,06:202 0 06,06:203 2 07,06:204 _ { 3 digits larger } look 06,01:203 0 { 06,01:203
0 + 1 = 1 } write 06,01:203 1 70,03:201 + 71,03:201 1 72,03:202 6 look 72,02:202 6 72,03:202 6 { 72,02:202 6
+ 72,03:202 6 = 1 2 } write 72,04:202 2 { carry the 1 } look 71,02:201 1 71,03:201 1 { 71,02:201 1 + 71,03:201
1 = 2 } { 2 + 1 = 3 } write 71,04:201 3 look 70,02:201 _ 70,03:201 + { 0 + 0 = 0 } { compare } look 70,04:201
_ 71,04:201 3 72,04:202 2 73,04:203 _ { 2 digits } look 04,06:201 1 05,06:202 0 06,06:203 2 07,06:204 _ { 3
digits larger } look 06,01:203 1 { 06,01:203 1 + 1 = 2 } write 06,01:203 2 70,05:201 + 71,05:201 1 72,05:202
6 look 72,04:202 2 72,05:202 6 { 72,04:202 2 + 72,05:202 6 = 8 } write 72,06:201 8 look 71,04:201 3 71,05:201
1 { 71,04:201 3 + 71,05:201 1 = 4 } write 71,06:201 4 look 70,04:201 _ 70,05:201 + { 0 + 0 = 0 } { compare }
look 70,06:201 _ 71,06:201 4 72,06:202 8 73,06:203 _ { 2 digits } look 04,06:201 1 05,06:202 0 06,06:203 2
07,06:204 _ { 3 digits larger } look 06,01:203 2 { 06,01:203 2 + 1 = 3 } write 06,01:203 3 70,07:201 +
71,07:201 1 72,07:202 6 look 72,06:202 8 72,07:202 6 { 72,06:202 8 + 72,07:202 6 = 1 4 } write 72,08:201 4 {
carry the 1 } look 71,06:201 4 71,07:201 1 { 71,06:201 4 + 71,07:201 1 = 5 } { 5 + 1 = 6 } write 71,08:201 6
look 70,06:201 _ 70,07:201 + { 0 + 0 = 0 } { compare } look 70,08:201 _ 71,08:201 6 72,08:202 4 73,08:203 _
{ 2 digits } look 04,06:201 1 05,06:202 0 06,06:203 2 07,06:204 _ { 3 digits larger } look 06,01:203 3 {
06,01:203 3 + 1 = 4 } write 06,01:203 4 70,09:201 + 71,09:201 1 72,09:202 6 look 72,08:202 4 72,09:202 6 {
72,08:202 4 + 72,09:202 6 = 1 0 } write 72,10:201 0 { carry the 1 } look 71,08:201 6 71,09:201 1 { 71,08:201
6 + 71,09:201 1 = 7 } { 7 + 1 = 8 } write 71,10:201 8 look 70,08:201 _ 70,09:201 + { 0 + 0 = 0 } { compare }
look 70,10:201 _ 71,10:201 8 72,10:202 0 73,10:203 _ { 2 digits } look 04,06:201 1 05,06:202 0 06,06:203 2
07,06:204 _ { 3 digits larger } look 06,01:203 4 { 06,01:203 4 + 1 = 5 } write 06,01:203 5 70,11:201 +
71,11:201 1 72,11:202 6 look 72,10:202 0 72,11:202 6 { 72,10:202 0 + 72,11:202 6 = 6 } write 72,12:201 6 look
71,10:201 8 71,11:201 1 { 71,10:201 8 + 71,11:201 1 = 9 } write 71,12:201 9 look 70,10:201 _ 70,11:201 + { 0
+ 0 = 0 } { compare } look 70,12:201 _ 71,12:201 9 72,12:202 6 73,12:203 _ { 2 digits } look 04,06:201 1
05,06:202 0 06,06:203 2 07,06:204 _ { 3 digits larger } look 06,01:203 5 { 06,01:203 5 + 1 = 6 } write
06,01:203 6 70,13:201 + 71,13:201 1 72,13:202 6 look 72,12:202 6 72,13:202 6 { 72,12:202 6 + 72,13:202 6 = 1
2 } write 72,14:201 2 { carry the 1 } look 71,12:201 9 71,13:201 1 { 71,12:201 9 + 71,13:201 1 = 1 0 } { 1 0
+ 1 = 1 1 } write 71,14:201 1 { carry the 1 } look 70,12:201 _ 70,13:201 + { 0 + 0 = 0 } { 0 + 1 = 1 } write
70,14:201 1 { compare } look 70,14:201 1 71,14:202 1 72,14:203 2 73,14:204 _ { 3 digits } look 04,06:201 1
05,06:202 0 06,06:203 2 07,06:204 _ { 3 digits equal } look 70,14:201 1 look 04,06:201 1 { 1 , 1 equal } look
71,14:202 1 look 05,06:202 0 { 1 , 0 smaller } look 70,12:201 _ 71,12:201 9 72,12:202 6 73,12:203 _ write
03,07:201 - write 05,07:201 9 06,07:202 6 look 06,06:203 2 look 06,07:202 6 { 2 , 6 larger } { borrow a 1 }
look 05,06:202 0 { change 0 to 9 } write 05,06:202 9 look 04,06:201 1 { 1 - 1 = 0 } write 04,06:201 0 { 1 2
- 6 = 6 } write 06,08:201 6 look 05,06:202 9 look 05,07:201 9 { 9 , 9 equal } { 9 - 9 = 0 } write 05,08:201
0 look 04,06:201 0 look 04,07:201 _ { 0 , 0 equal } { 0 - 0 = 0 } write 04,08:201 0 look 04,08:201 0 write
04,08:201 _ look 05,08:201 0 write 05,08:201 _ look 06,08:201 6 { read the answer } look 06,08:201 6 07,08:202
_ look 07,02:205 _ { final remainder is 6 }
```

## 4.2 Evaluation

Within the training set, strings such as those in Figure 3 can be interpreted as sequences of actions performed by a demonstrator in response to an input question. At test time, new input questions of the form "What is the remainder when __ is divided by __?" or "Calculate the remainder when __ is divided by __" were presented to the model, which then had to generate continuations; the continuations the model generated can be interpreted as the sequences of actions that it 'expected' to follow the test questions. Continuations were evaluated as correct if they ended in a no-op of the form "{ final remainder is $X$ }" where $X$ was the correct answer, and as incorrect otherwise. As the implementation of GPT-Neo we used has a maximum generation length, some generations did not include an end-of-sequence token; in such cases, the generation was fed back to the model as a prefix (with the first 500 tokens removed from the beginning, as there is also a maximum window size), with this process repeating up to a maximum of 25 times until an end-of-sequence token was generated.

Interpreting the tokens generated by the model as actions to be taken in the environment is straightforward for actions that can always be executed. For example, a well-formed string of the form "write *coord w*" can always be interpreted as the action of writing the symbol $w$ at the x-y coordinate *coord*; if *coord* already contains a symbol, it is overwritten. In contrast, in the demonstrations provided at training, the symbol $s$ at the end of strings of the form "look *coord s*" is constrained by the environment; it represents the symbol that the demonstrator saw when observing x-y coordinate *coord*. However, at inference time, a model could theoretically generate a symbol sequence such as "write $coord_1$ 5 look $coord_1$ 8"; interpreting this as a sequence of actions is nonsensical, as it implies that the model wrote a 5 at a particular x-y coordinate, then immediately looked at the same coordinate and saw an 8 there. To address this, a separate script monitored model output as it was generated and attempted to interpret the model output as a sequence of actions. The initial intent was for this process to halt generation immediately whenever the model produced a string of the form "look *coord*", at which point it would append the symbol currently located at *coord* in the environment, and restart generation with this amended output used as the prefix. In practice, it was most efficient only to do this only in the rare cases when the symbol that the model generated was different from the symbol that it had previously "written" at the corresponding coordinate. We refer to this approach as *environment forcing* by analogy to related concepts such as professor forcing[33] and attention forcing[34], as it allows the state of the environment to impose constraints on the generated text, with the goal of mitigating exposure bias. It also allows information to be stored in the environment and later retrieved even if this information were to fall outside of the context window[2].

**Table 1**. Number of questions in the validation set answered correctly by each model, out of 100. The highest-scoring model was used for the final evaluations reported in Table 2. When two models scored equally well, the one that took fewer iterations to train was chosen.

| Actions | Iterations | | | | | | | | | | |
|---|---|---|---|---|---|---|---|---|---|---|---|
| | 5K | 6K | 7K | 8K | 9K | 10K | 11K | 12K | 13K | 14K | 15K |
| Writing, looking, no-ops | 84 | 71 | 78 | 81 | 85 | **86** | 80 | 77 | 86 | 79 | 84 |
| Writing, looking | 63 | 57 | 65 | 59 | 59 | 58 | 71 | 70 | 65 | **75** | 65 |
| Writing | 2 | 2 | 1 | 2 | **4** | 1 | 1 | 0 | 1 | 1 | 4 |
| None | 0 | 0 | 1 | **3** | 1 | 2 | 1 | 2 | 2 | 3 | 2 |

---

[2] As noted earlier, GPT-Neo's maximum window size meant that actions generated at the beginning of an action sequence were not necessarily guaranteed to be accessible later in the action sequence. Environment forcing provides a workaround by allowing the environment to 'remind' the model what symbol exists at some location whenever the model generates a directive to look at that grid location. This is analogous to how students solving mathematical problems by hand make use of pencil and paper, periodically glancing at relevant parts of their work so as not to have to hold the entire problem in memory at all times.

## 5  Results

Table 2 reports the percentage of questions in the test sets answered correctly by each model. In addition, we conducted a manual error analysis of the strings generated by the best-performing model for the 91/500 questions on the interpolated test set that it answered incorrectly, to identify the nature of the first error committed by the model in each case. The most common were simple transcription errors: Forty-four of the incorrectly answered problems had been computed exactly correctly until the final step when the final answer had to be rendered into the form "{ final remainder is __ }", with the answer often being mis-transcribed by a single digit. In another thirteen problems, the first error occurred when copying a number from one part of the page to another, e.g. from a "scratch" area to the division problem proper. In twelve, the problem was mistranscribed from the get-go, with the error occurring in the very first action. Finally, twelve errors occurred during numeric comparisons, six while completing subtraction operations, and there were four other errors generally caused by a failure to complete some necessary action, such as bringing a digit down from the dividend.

**Table 2**. Percentage of questions in the test sets answered correctly by each model.

| Actions included in generations | Interpolated test set performance | Train-easy/medium/hard combination test set performance |
| --- | --- | --- |
| None | 1.8% | 2.6% |
| Writing | 1.0% | 2.8% |
| Writing, looking | 60.2% | 66.8% |
| Writing, looking, no-ops | 81.8% | 85.4% |

*Note.* A no-op of the form "{ final remainder is $X$ }" appeared at the end of each demonstration in each condition.

**Figure 4.** Error analysis of the strings generated by the best-performing model, for the 91 questions on the interpolated test set that it answered incorrectly.

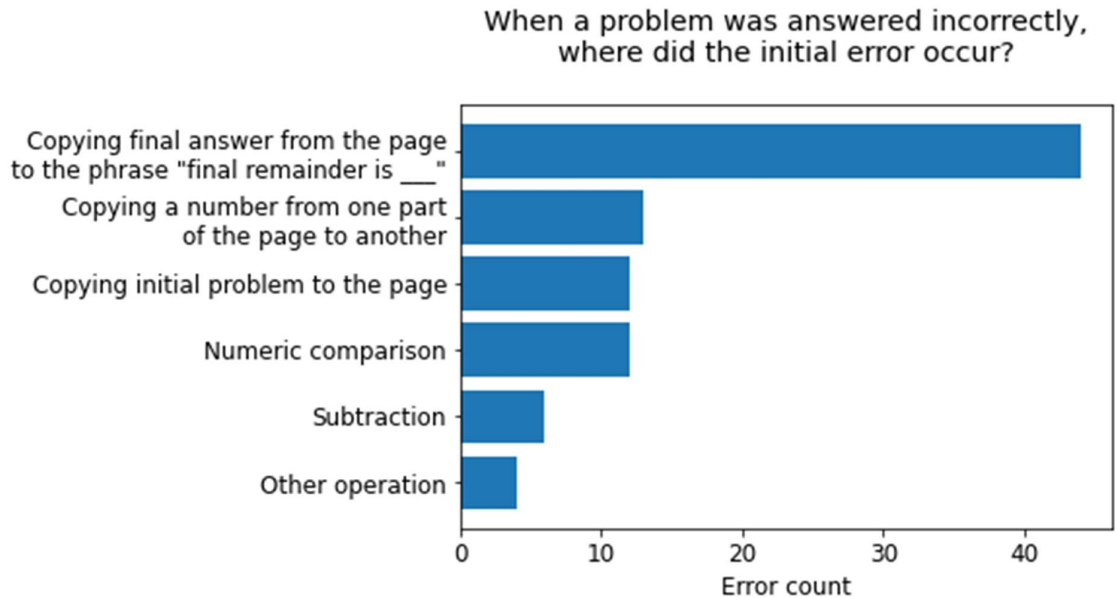

## 6 Discussion

Unsurprisingly, with only two hundred training examples, fine-tuning on a dataset that only contained prompts followed immediately by answers led to extremely poor performance; the same was true of a dataset that included demonstration containing *write* actions only. However, including *look* actions with environment forcing improved performance substantially, and the addition of strategically placed no-ops produced further gains. Some intuition for why the inclusion of *look* actions led to improved performance can be gained by considering an example. One of the first key actions in the long division algorithm as described by [36] is to determine which is smaller, the divisor or the number consisting of the first *n* digits in the dividend, where *n* is the number of digits in the divisor; the next step of the problem differs depending on whether the divisor is larger or smaller than this number. This is a relatively difficult pattern to divine without any hint that the first *n* digits of the dividend are the relevant ones, and that they should be compared to the digits of the divisor. By providing *look* actions, however, the pattern decomposes into several simpler ones that are easier for gradient descent to pick up on. In particular, we look at the first digit of the dividend, then look at the first digit of the divisor. If the first digit of the dividend is larger or smaller than the divisor, we are done; if the digits are equal, we repeat the process but looking at the second digits rather than the first, and so on. Because comparison operations are conducted frequently in the course of a single division problem, even after only two hundred training examples, the model has seen many cases in which "*look coord$_1$ s$_1$*" is followed by "*look coord$_2$ s$_2$*", and in which the symbol that follows differs depending on whether $s_1$ is less than $s_2$, or $s_1$ is greater than $s_2$, or $s_1$ and $s_2$ are the same digit. For example, if $s_1$ is less than $s_2$, then this means the divisor is smaller and that we are ready to determine how many times it goes into the leftmost *n* digits of the dividend, which in turn means that the next symbols to appear will be of the form *write coord$_3$ 0 clear*.

In other words, the *look* actions help to make the relevant steps of the problem explicit, likely facilitating shaping (cf. [38]). In this particular case, they also enforce a more consistent pattern: irrespective of the number of digits in the divisor, the next symbols will be of the form *write coord$_3$ 0 clear* if and only if the preceding symbols fit the pattern *look coord$_1$ s$_1$ look coord$_2$ s$_2$* where $s_1$ is less than $s_2$.

Note that this explanation of the benefits provided by "*look*" actions also helps explain the benefits that were derived from including appropriate no-ops. For example, a substring of the form "*look coord$_1$ s$_1$*" is followed by "*look coord$_2$ s$_2$*" in the context of subtraction operations as well as comparison operations, and the actions that should follow differ in these two contexts. There are a number of subtle contextual cues that can help the model discriminate between these contexts, but regularly providing a no-op that contains a consistent set of symbols when we are in a comparison context, and not providing that no-op when we are in a subtraction context, may make the discrimination task easier. While the no-ops we used consisted only of a few words, in some contexts full natural language explanations may provide usefully relevant information [39].

Hendrycks et al [19] discuss the value of step-by-step solutions in a context where a model pre-trained on a diverse 23GB corpus of mathematical problems was fine-tuned on a dataset of 7,500 problems from mathematics competitions. During fine-tuning, they trained the model on an equal mix of instances having the form "<Problem> Final Answer: <Answer>" and "<Problem> Full Solution: <Step-by-step solution>". At inference time, this allowed the model to be prompted either to generate either full step-by-step solutions or final answers only. In contrast to the present approach, which provided sequences of very low-level actions intentionally structured to make relevant patterns easier to detect, the steps of their step-by-step solutions in were substantially higher-level, e.g., a handful of sentences consisting of a mix of natural language and mathematical notation. They found that the model was less likely to arrive at correct answers when prompted to produce full solutions vs. final answers only, suggesting that the model was unable to leverage its own generated step-by-step solutions effectively. However, this did not mean that it was unable to leverage step-by-step solutions at all, as they found its performance improved when it was provided with partial solutions; they also found that including step-by-step solutions during training increased accuracy. They summarized these findings as follows: "Our results show that models can make use of actual step-by-step solutions provided to them in various ways, but that they are still unable to effectively use their own generated solutions. Bridging this gap poses an interesting direction for further research" ([19], p. 2). The present study adds one girder to this bridge by providing a case study of a model using its own generated step-by-step action sequences to arrive at correct answers, and doing so much more effectively than when these step-by-step examples are not provided at training time.

### 6.1 Interpretability

It is sometimes claimed that there is a tradeoff between interpretability and performance, such that making a model more performant requires introducing complexity that leads to less interpretable models. Interestingly, the fact that the present approach generates not only answers but a step-by-step accounting of how they were arrived at means that it provides a limited sort of post hoc interpretability[40] that can be exploited to improve subsequent models. Furthermore, the improvements frequently involve spelling out the sequence of actions in the training demonstrations in greater detail. This higher level of detail is then reflected in the outputs produced during validation, arguably making the model more interpretable, not less. For example, in an earlier version of the schema illustrated in Figure 3 for representing the steps of a long division problem, examination of the output revealed that it was common for the model to err when comparing which of two multi-digit numbers was larger. Changes were made to the input to make the steps involved in comparing two multi-digit numbers more explicit—e.g., including "look" actions focusing on the relevant digits—and performance improved. Furthermore, these added "look" actions make it possible to identify commonalities in the handful of multi-digit comparisons that the model still gets wrong: several of these involved comparisons involving very long (6- or 7-digit) numbers, which may not have been well represented in the training data. The error analysis in Figure 4 suggests that there is additional room for improvement elsewhere as well. Specifically, the current results suggest that performance could be improved substantially by tweaks that reduced the number of simple transcription errors, whereby the answer was copied incorrectly from the page into the final answer, or from the initial prompt onto the page, or from one part of the page to another. These errors often involved numbers that were copied from earlier in the demonstration string being swapped, duplicated, or omitted, suggesting that the use of alternative positional encodings could be a promising approach [20], [41]. Ultimately, when a model is forced to "show its work" by generating a full list of actions or reasoning steps that led to its final result, stumbling blocks are easier to identify and address.

## 6.2 Working memory

In humans, 'working memory' refers to a cognitive system with a limited capacity that "provides temporary storage and manipulation of the information necessary for such complex cognitive tasks as language comprehension, learning, and reasoning" ([42], p. 556). It includes constructs such as the "visuospatial sketchpad"—identified with the ability to represent visuospatial information in the 'mind's eye'—and an analogous "phonological loop" for auditory information. Both systems have a limited capacity: for a vivid illustration of the limited capacity of the phonological loop, contrast the ease of remembering seven random digits by repeating them to oneself multiple times with the difficulty of doing the same with fourteen random digits. Although we have described the symbols generated at evaluation time as if they describe a sequence of actions, there may also be some connection to the function served by working memory. Specifically, the generated stream serves as a kind of 'workspace' in which symbols can be stored and referred to in the future, and in which diffuse patterns can be reified as single symbols, which can in turn participate in future operations. In addition, the use of environment forcing allows the environment to serve as a kind of external memory, permitting symbols to be stored and brought back into the context window when required simply by 'looking' at the appropriate coordinate.

Some overarching theories of cognitive function, such as global workspace theory and 'neural blackboards', likewise hypothesize that a central shared resource for storing and manipulating visual and auditory content is important to cognitive tasks that require complex symbolic processing. Although a Transformer's context window is normally not conceived of as a resource that the model 'manipulates,' during inference its contents are certainly affected by the symbols that the model itself has recently generated. Thinking of the context window as a potential storage location for temporary, intermediate representations of data may prime intuitions for why the generation of appropriately crafted step-by-step solutions can be useful to the model.

## 6.3 Limitations

A downside of this approach is that a bespoke training set may need to be developed for each task of interest; the fact that the datasets needed are small alleviates but does not eliminate this issue. Additionally, although we have noted that the algorithmically generated demonstrations in this study in principle could have been replaced with demonstrations manually generated by humans, the variability associated with such demonstrations could mean that the approach is less effective with human demonstrations in practice, or when applied to tasks with more inherent variability than long division. Furthermore, while one might hope for generalizability after fine-tuning on a sufficient number of reasoning and symbol manipulation tasks with a common 'language' or protocol for representing action and state information, this may or may not be observed[43]. The use of alternative demonstration representations that bear more similarity to natural language may also have been able to better leverage the language model's pretraining. Prompt-tuning [44], [45] may work particularly well with representations of this kind, potentially enabling more rapid adaptation to new tasks [46].

### 6.4 Summary of contribution

Although the potential of autoregressive models for behavior cloning-based imitation learning has been previously demonstrated, the present study offers a concrete illustration of several points. First, it suggests that it may be possible to train extant autoregressive language models on very small training sets consisting of fine-grained sequences of actions to make headway on tasks involving reasoning and complex symbol manipulation[3]. In particular, the inclusion of actions that communicated which parts of the environment the demonstrator was paying attention to ("look" actions) dramatically improved performance.

The fact that autoregressive models are so capable with respect to language-based tasks suggests that for some procedures, it may be helpful to provide text that communicates how to complete the fine-grained steps of a procedure that led to a final result in natural language. For symbolic manipulation tasks that have yet to be cracked by Transformer-based models, this could in principle be combined with the approach of providing fine-grained symbolic state and action information at training time illustrated here, e.g., by including natural language step-by-step "explanations" of how to get from the question to the solution within no-ops. There is certainly no guarantee that this would work in the general case. As mentioned previously, when Hendrycks et al [19] provided step-by-step natural language explanations of how to derive the answers to complex mathematical questions to a Transformer-based model during training, they found that although the provision of step-by-step solutions during training did increase accuracy, performance was worse when the model was prompted to provide step-by-step explanations rather than answers only. That said, the contrasting finding in this case study suggests that there are some circumstances in which step-by-step descriptions can be leveraged at inference time to improve performance on mathematical tasks. In addition to the fact that the "step-by-step explanations" provided by [19] were in English while the demonstrations in this study represented state and action information symbolically, the demonstrations in this study were also far more fine-grained and detailed. Future research may wish to test such models' performance when they are provided with sufficiently detailed demonstrations of the state and action information associated with each step, perhaps interleaved with relevant natural language explanations or rationales for the different steps.

Finally, the benefits of the post-hoc interpretability discussed in section 6.2 should not be underestimated. The high level of explicit detail (Figure 3) allows us to observe exactly where generations that lead to incorrect results go wrong (Figure 4). This in turn permits the tweaking of one's approach to demonstrating those parts of the task that are most difficult for the network to generate accurately, as described in section 6.2. However, it should be borne in mind that the step-by-step "demonstrations" generated by the model provide only a limited sort of interpretability, and should not be treated as true or accurate explanations of how the final answer was arrived at—even if the answer is correct, as an example provided by Hendrycks et al. clearly demonstrates ([19] sec 4.1). However, given that model-generated demonstrations will generally exert some influence over the final answer, their content may still help us understand more about what information the model was leveraging to achieve correct answers, and what steps the model is getting tripped up on in the case of errors.

### 6.5 Conclusions

In sum, this paper has presented a case study in which a small training set was used to fine-tune an otherwise unmodified 125M-parameter GPT-Neo model to complete the "remainder after division" task (i.e., the modulo operation), among the most difficult of the tasks in the DeepMind Mathematics Dataset for Transformer-based models. Test-set performance of 82%-85% was achieved with a set of two hundred training examples[4] that included step-by-step demonstrations of correctly solved long division problems expressed in a symbolic format. Training sets that only contained questions and answers with no demonstrations led to far worse performance, as did training sets that included less complete demonstrations. This illustrates that at least in some cases, providing step-by-step demonstrations of the path from a question to the solution can enable Transformers to make use of their own generations to achieve high accuracy on symbolic manipulation tasks, although careful curation of training data may be needed. For incorrectly answered questions in validation data, the generated demonstrations can be inspected to help inform where the model is going off the rails, which is useful in illustrating what aspects of the demonstrations in the training data require alternate representations or fuller detail. It is encouraging that minimal training data is required, as it suggests that with appropriate tooling (i.e., software that makes it easy to provide demonstrations of various kinds by pointing and clicking), people without expertise in machine learning may be able to enable general-purpose models to perform a wider variety of tasks by participating in the development of appropriate training sets. More broadly, this approach fits into the concept of "data-centric" approaches to achieving increased performance by making iterative improvements to training data rather than to learning algorithms, recently endorsed by some experts[11], [47].

---

[3] For example, Transformers currently have difficulty with "grouping games"[44], a form of logic problem associated with law school admission tests, but it seems plausible that performance could be improved substantially if training proceeded on examples that explicitly laid out the sequences of steps that a human might take to solve each problem, perhaps with the aid of an environment similar to the one used in the present paper to keep track of the relationships between the relevant entities. Substantially improved performance on benchmarks with relatively narrowly defined tasks like AR-LSAT[44], or the few problem sets from the DeepMind Mathematics Dataset[8] that have proven difficult for Transformers to crack (e.g. change of base, listing prime factors), are therefore conceivable with the techniques used here.

[4] As a validation set of 100 examples was also used to select the number of iterations that it was optimal to train the model for, it could be argued that the information in a total of three hundred examples contributed to the final model selection. That said, most reasonable choices for 'number of iterations' tested yield comparable performance (Table 1).

## 7 Acknowledgments

## 8 Data availability

Training, validation, and test data and the code for generating it is available at https://github.com/mesotron/teaching_transformers. Code for fine-tuning GPT-Neo and evaluating the results can be found at https://colab.research.google.com/drive/1glgRxBepDVz6Lw2_cnsWbL6xZJXAthY3.